\documentclass[sigconf]{acmart}
\AtBeginDocument{%
  }

\setcopyright{acmlicensed}
\copyrightyear{2025}
\acmYear{2025}
\setcopyright{acmlicensed}
\acmConference[MM '25]{Proceedings of the 33rd ACM International Conference on Multimedia}{October 27--31, 2025}{Dublin, Ireland.}
\acmBooktitle{Proceedings of the 33rd ACM International Conference on Multimedia (MM '25), October 27--31, 2025, Dublin, Ireland}
\acmISBN{979-8-4007-2035-2/2025/10}
\acmDOI{10.1145/3746027.3755085} 
\usepackage{multirow} 

\begin{document}

\title{
DMC$^3$: Dual-Modal Counterfactual Contrastive Construction for Egocentric Video Question Answering
}
 \author{Jiayi Zou}
 \orcid{0009-0003-8821-6358}
 \affiliation{
   \institution{Nanjing University of Posts and Telecommunications}
   \city{Nanjing}
   \country{China}
 }
 \email{2023010213@njupt.edu.cn}
 
 \author{Chaofan Chen}
 \orcid{0000-0001-7970-7698}
\authornote{Corresponding author.}
 \affiliation{
   \institution{Institute of Automation, Chinese Academy of Sciences }
   \city{Beijing}
   \country{China}
 }
 \email{chencfbupt@gmail.com}

 \author{Bing-Kun Bao}
 \orcid{0000-0002-1809-5592}
 \authornotemark[1]
 \affiliation{
   \institution{Nanjing University of Posts and Telecommunications}
   \city{Nanjing}
   \country{China}
 }
  \affiliation{
   \institution{Peng Cheng Laboratory
   }
   \city{Shenzhen}
   \country{China}
 }
 \email{bingkunbao@njupt.edu.cn}

 \author{Changsheng Xu}
 \orcid{0000-0001-8343-9665}
 \affiliation{
   \institution{MAIS, Institute of Automation, Chinese Academy of Sciences, \\
   }
   \city{Beijing}
   \country{China}
 }
 \affiliation{
   \institution{Peng Cheng Laboratory
   }
   \city{Shenzhen}
   \country{China}
 }
 
 \email{csxu@nlpr.ia.ac.cn}



\begin{abstract}

Egocentric Video Question Answering plays an important role in egocentric video understanding, which refers to answering questions based on first-person videos. Although existing methods have made progress through the paradigm of pre-training and fine-tuning, they ignore the unique challenges posed by the first-person perspective, such as understanding multiple events and recognizing hand-object interactions. To deal with these challenges, we propose a Dual-Modal Counterfactual Contrastive Construction (DMC$^3$) framework, which contains an egocentric videoqa baseline, a counterfactual sample construction module and a counterfactual sample-involved contrastive optimization. Specifically, We first develop a counterfactual sample construction module to generate positive and negative samples for textual and visual modalities through event description paraphrasing and core interaction mining, respectively. Then, We feed these samples together with the original samples into the baseline. Finally, in the counterfactual sample-involved contrastive optimization module, we apply contrastive loss to minimize the distance between the original sample features and the positive sample features, while maximizing the distance from the negative samples. Experiments show that our method achieve 52.51\% and 46.04\%  on the \textit{normal} and \textit{indirect} splits of EgoTaskQA, and 13.2\% on QAEGO4D, both reaching the state-of-the-art performance.
The code is available at https://github.com/trea1262/DMC\_3.
\end{abstract}





\maketitle

\section{Introduction}

\begin{figure}
    \centering
    \includegraphics[width=1.0\linewidth]{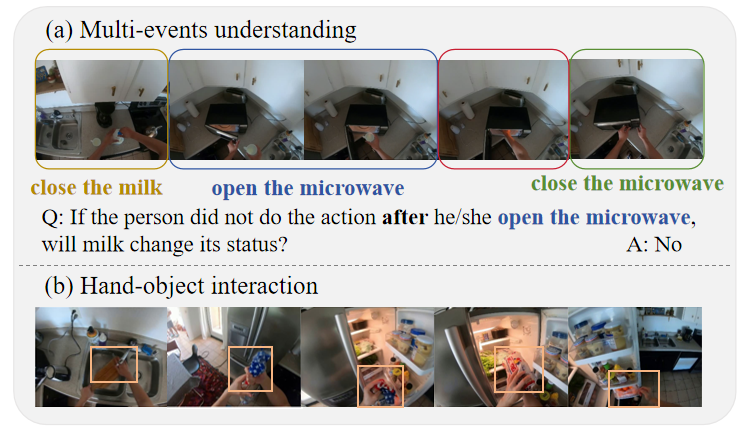}
    \caption{Two challenges of Egocentric VideoQA. 
    (a) Egocentric video has a long time span and contains multiple fine-grained events, requiring the model to understand these events and the contextual information.
    (b) Egocentric video emphasizes hand-object interactions, requiring the model to identify and interpret relevant visual regions.
}
    \label{fig: intro}
\end{figure}

As Embodied Artificial Intelligence (EAI)~\cite{majumdar2023we,ren2024embodied} becomes more integrated into daily life, the study of egocentric video understanding~\cite{wang2023all,phan2024henasy} has grown increasingly prominent.
Egocentric Video Question Answering (Egocentric VideoQA)~\cite{di2024grounded,zhou2025egotextvqa} is an important branch of this research line,
which aims to answer questions about the events recorded from the first-person perspective.
Research on this task can help enhance augmented and virtual reality (AR/VR) devices, offering deeper insights into personal visual experiences and interactions~\cite{mu2024embodiedgpt,plizzari2024outlook}.

To tackle the Egocentric VideoQA task, we need to solve two major challenges.
\textbf{(1)} Firstly, the model needs to have strong multi-event and contextual understanding capabilities. 
Compared to exocentric (third-person) videos, 
egocentric videos contain a large number of events and provide fine-grained segmentation of them\cite{chen2024egocentric,huang2024egoexolearn}.
As illustrated in Figure~\ref{fig: intro}(a), an egocentric video of heating milk is divided into multiple fine-grained actions, such as "close the milk" and "open the microwave".
Furthermore, the questions typically include temporal descriptors like "after", which requires the model to further comprehend the contextual information between events
during the question-answering process. 
\textbf{(2)} Secondly, egocentric videos emphasize interactions between the wearer's hands and objects.
As depicted in Figure~\ref{fig: intro} (b), almost every frame in the dataset captures the hand-object interactions, and most of them appear in the center of the frame. 
This requires the model to prioritize the interactions occurring in the center while disregarding warped regions at the edge of the viewing area, as these areas may introduce noise.
In summary, 
methods designed for Exocentric VideoQA~\cite{xiao2022video,jin2023knowledge,min2024morevqa}  are not suitable to 
address the above challenges because the egocentric videos have unique perspective and focus, requiring specially designed datasets and approaches~\cite{wang2023ego,nagarajan2023egoenv}.

Recently, researchers have proposed datasets such as EgoVQA~\cite{fan2019egovqa}, EgoTaskQA~\cite{jia2022egotaskqa} and QAEGO4D~\cite{barmann2022did}, laying the foundation for research in Egocentric VideoQA. 
On this basis, previous studies~\cite{lei2021less,pramanick2023egovlpv2,pei2024egovideo} 
mainly focus on developing Vision-and-Language Pre-training (VLP) frameworks. 
%
%
%
These frameworks leverage large-scale datasets, such as EGO4D\cite{grauman2022ego4d}, for self-supervised pre-training. 
At this stage, they encode video actions and corresponding text descriptions into a shared embedding space, 
establishing correspondence between visual and textual information in the first-person perspective. 
However, when subsequently fine-tuned for VideoQA, they often struggle with complex scenes that involve multiple events or interactions,
leading to sub-optimal results. 
%
%
%
%
%
%
To address the challenges in Egocentric VideoQA, recent approaches have explored diverse strategies.
One category enhances comprehension through specialized visual and language interaction mechanisms~\cite{zhangmulti,zou2024language}, such as partitioning video frames and utilizing predefined matrices to model interactions between elements. 
%
%
%
%
Another category leverages video grounding to support question answering, integrating video segments~\cite{di2024grounded} or identifying relevant time intervals~\cite{chen2024groundedmultihopvideoqalongform} to guide response generation.
%
%
%
However, these methods are often computationally expensive and rely on external information to ensure the effectiveness of the reasoning process.
%
%
%
%

In this work, we propose 
a novel Dual-Modal Counterfactual Contrastive Construction (DMC$^3$) framework, 
which generates positive factual and negative counterfactual samples across two modalities for 
contrastive learning to answer questions.
Specifically, the proposed DMC$^3$ contains an egocentric videoqa baseline, a counterfactual sample construction module and a counterfactual sample-involved contrastive optimization module.
\textbf{(1)} In the baseline, we formulate Egocentric VideoQA as a classification task. A dual-stream encoder is employed to extract and fuse the features of videos and questions.
\textbf{(2)} In the counterfactual sample construction module, we construct paired factual and counterfactual samples for visual and textual modalities respectively, and treat them as positive and negative instances. 
%
%
In the textual modality, we propose a novel strategy called Event Description Paraphrasing (EDP). 
For positive samples, EDP produces variants of the original question through synonym substitution, introducing subtle linguistic shifts while preserving the semantic intent. 
For negative samples, EDP replaces the temporal references, such as substituting "first" with "last", along with masking events to obscure specific contextual cues.
%
%
In the visual modality, we introduce a complementary strategy called Core Interaction Mining (CIM). 
For positive examples, CIM generates enhanced samples by diversely retaining areas of the video frame that emphasize interactions
between hands and objects 
while masking other areas. 
For negative examples, CIM reverses the positive approach by masking the interaction-centric areas and retaining the previously masked areas.
%
%
%
%
%
%
\textbf{(3)} In the counterfactual sample-involved contrastive optimization module, 
we encourage the model to bring its fused features closer to those of positive samples while 
keeping them from negative samples. 
In this way, 
positive samples generated by EDP
assist in mining semantically consistent information from different contexts,
while negative samples 
enable the model to concentrate on crucial content, such as event descriptions and temporal aspects.
Additionally, modifying time descriptions in negative samples further enhances the 
model to distinguish adjacent events
and allows for the tracking of event flows.
In the visual modality, 
the positive samples generated by the CIM help the model capture key interaction information from the frame representation.
Meanwhile, the negative samples enhance the modeling of hand-object interactions by indicating what constitutes noise, allowing the model to refine its focus.
%
Therefore, based on the above analysis, our proposed framework can enhance the model's ability to understand multiple events and hand-object interactions.
%
%
%
%


In summary, our contributions can be outlined as follows.
\begin{itemize}
    \item To tackle the challenges of multi-event understanding and hand-object interactions, we propose a Dual-Modal Counterfactual Contrastive Construction (DMC$^3$) framework, which generates positive and negative samples of videos and questions for contrastive learning.
    \item Experiments demonstrate that our approach has achieved state-of-the-art performance on the \textit{normal} and \textit{indirect} splits of EgoTaskQA and QAEGO4D. Ablation studies confirm the effectiveness of every component.
\end{itemize}

\section{Related work}

In this section, we review the related work on Egocentric VideoQA, 
contrastive learning approaches and counterfactual sample generation approaches.

\subsection{Egocentric Video Question Answering}

With the increasing application of intelligent wearable devices, current research has begun to focus more on egocentric video understanding and the interaction between these devices and humans~\cite{zhang2023helping,dai2024gpt4ego,salehi2024actionatlas,liu2023advancing}. 
Previous researchers have proposed datasets on egocentric video understanding and video question answering in this field. 
EgoVQA~\cite{fan2019egovqa} was among the first to bring VideoQA into the domain of egocentric video understanding. 
Following this, EgoTaskQA~\cite{jia2022egotaskqa} was developed to design questions and answers related to events in egocentric videos.
Additionally, datasets for egocentric video understanding have laid the foundation for enriching VideoQA datasets. 
EPIC-KITCHENS-100~\cite{damen2022epic} is a dataset that focuses on kitchen scenes, collecting 100 hours egocentric videos in this setting. 
Ego4D~\cite{grauman2022ego4d} has amassed over 3,600 hours of egocentric video data, with annotations defining tasks such as egocentric action recognition, object detection, hand-object interactions and visual query grounding. 
Building upon these video understanding datasets, QAEGO4D~\cite{barmann2022did} created corresponding questions and answers to further enhance the field of VideoQA.
Existing methods~\cite{zhao2023learning,wang2024videoagent}
primarily focus on building Vision-Language Pre-training (VLP) frameworks to pre-train models on egocentric video understanding datasets. This process aligns visual and textual modal features, followed by fine-tuning for VideoQA.
A dual-stream encoder pre-training framework, EgoVLP~\cite{lin2022egocentric}, was designed to align the features of the two modalities. 
EgoVLP achieved better predictive performance on first-person VideoQA datasets compared to exocentric VideoQA methods~\cite{jiang2020reasoning,le2021hierarchical}. 
Building upon this work, EgoVLPv2~\cite{pramanick2023egovlpv2} enhances the model performance by integrating cross-attention mechanisms within the encoders. 
This allows for more effective fusion between the visual and textual streams, but still struggles with challenges specific to Egocentric VideoQA. 
On the one hand, some methods improved performance by enhancing the visual and text interaction mechanism.
Zou \textit{et al.}~\cite{zou2024language} introduced a language-aware gating mechanism to replace the traditional cross-attention mechanism, and designed sparse sampling and visual refinement modules to enhance feature learning.
%
%
MFAS~\cite{zhangmulti} built upon EgoVLPv2 and integrated a patch partitioning and merging module, a prior-guided patch selection module, and a hierarchical aggregation network. 
%
On the other hand, some methods introduced video grounding~\cite{chen2024groundedmultihopvideoqalongform,di2024grounded} to assist question answering.
However, these methods increase computational demands and introduce prior knowledge.
In contrast, our method utilizes fewer parameters and does not rely on additional prior knowledge.

\subsection{Contrastive Learning Approaches}

Contrastive learning~\cite{tian2020makes,liu2023multimodal}
has been widely applied to multi-modal tasks such as Visual Question Answering (VQA)~\cite{liu2023enhancing,10814063,10310122} and Visual Commonsense Reasoning (VCR)~\cite{li2023vision,10888014}, achieving highly effective results. 
The core idea is to generate positive samples through data augmentation techniques based on given samples, while negative samples are derived from other samples or by altering the original sample. 
Contrastive loss is then applied to minimize the distance between the features of the original and positive samples, while maximizing the distance between the features of the original and negative samples.

Chen \textit{et al.} proposed the CSS~\cite{chen2020counterfactual} and CSST~\cite{chen2023counterfactual} models, which enhance VQA by generating positive and negative sample pairs through modifications to both images and textual descriptions. 
CSS mitigates language bias by introducing a branch of question-answering based on the analysis of causal reasoning.
On this basis, CSST leverages contrastive learning to improve the alignment between visual and textual representations, thereby enhancing the performance of multi-modal understanding tasks. 
%
Zhang \textit{et al.}~\cite{zhang2021multi} introduced contrastive learning into VCR.
They constructed informative positive and negative contrast samples at the image, object, and text levels within a batch, which helped to efficiently extract discriminative representations for VCR.

\subsection{Counterfactual Sample Generation Approaches}
Counterfactual sample generation is a powerful data augmentation method that has been extensively evaluated in visual understanding tasks, such as image captioning~\cite{guo2020nonautoregressiveimagecaptioningcounterfactualscritical} and visual question answering~\cite{wen-etal-2023-digging}. 
Initially, this method focuses on expanding datasets by enhancing image content and textual descriptions to improve model performance~\cite{wei2019eda,falcon2022feature,zhao2023learning}. 
For instance, Mashrur \textit{et al.}~\cite{mashrur2024robust} generated multiple semantically-consistent but heterogeneous instances from the visual and textual inputs, which were then fed into the model and the predictions were combined for a more robust output.
Zhang \textit{et al.}~\cite{zhang2024if} evaluated the counterfactual reasoning abilities of contemporary multi-modal language models by constructing counterfactual samples within the questions of a visual question answering.

In our work, we introduce counterfactual sample generation specifically tailored for Egocentric VideoQA. 
Our approach leverages the dynamic nature of video data, allowing for the generation of semantically consistent and contextually rich counterfactual samples. 
Unlike traditional methods that may focus solely on static images or text, our technique integrates both visual and textual modalities, providing a comprehensive augmentation strategy. 
%

\section{Method}

\begin{figure*}
    \centering
    \includegraphics[width = 1\linewidth]{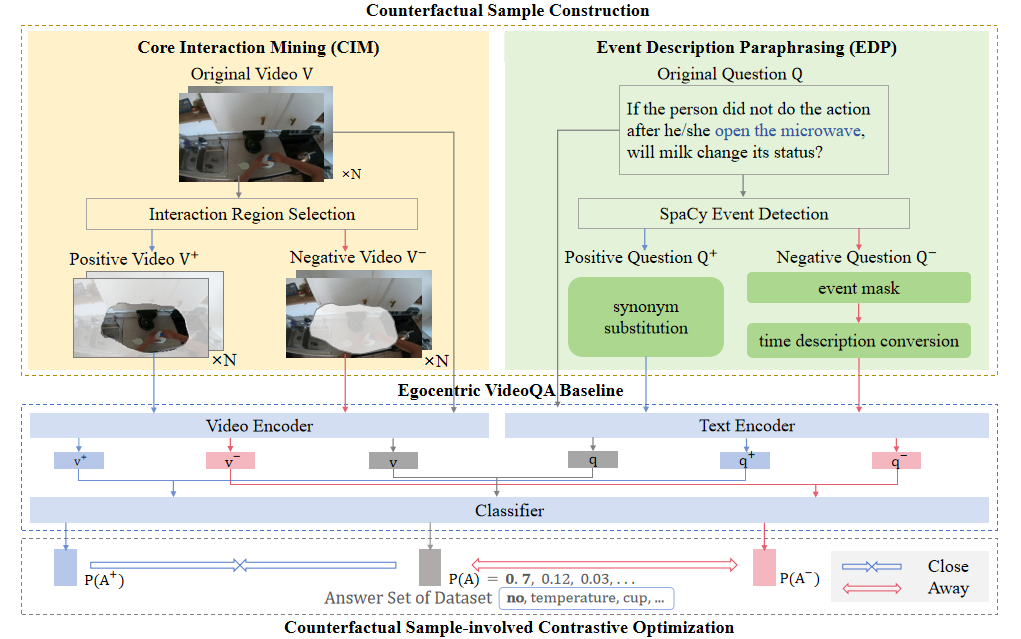}
    \caption{The 
    Dual-Modal Counterfactual Contrastive Construction framework comprises a egocentric videoqa baseline, a counterfactual sample construction module and a counterfactual sample-involved contrastive optimization module. 
    CIM and EDP are utilized to generate factual and counterfactual samples for visual and textual modalities, respectively. 
    Positive factual samples are integrated into the model through paths denoted by blue arrows, while the paths of negative counterfactual samples are represented by red arrows.
}
    \label{fig:2}
\end{figure*}

In this section, we introduce the technical details of the proposed
Dual-Modal Counterfactual Contrastive Construction 
(DMC$^3$) framework.
As illustrated in Figure~\ref{fig:2},
DMC$^3$ comprises an egocentric videoqa baseline, a counterfactual sample construction module 
and a counterfactual sample-involved contrastive optimization module.
%
Our model undergoes a two-stage training process, which will be detailed below.

\subsection{Egocentric VideoQA Baseline}
Our egocentric videoqa baseline is a dual-stream encoder structure consisting of a visual encoder, a text encoder, and a classifier.

Given a video $V$,
we first randomly extract N frames and use the 
encoder of the TimeSformer~\cite{bertasius2021space} to extract video features $v\in \mathbb {R}^{N \times C\times H\times W}$, where $C$, $H$ and $W$ represent channels, height, and width, respectively. The text encoder is a 12-layer Transformer~\cite{han2022survey}, which encodes the input question into the representation $q\in \mathbb {R}^{l \times d}$, where $l$ is the length of token sequence and $d$ is hidden dimension. 
In each decoder layer of TimeSformer, we employ the cross-attention mechanism to fuse the video (keys and values) and text representations (queries).
%
Finally, we feed the fused features into
the classifier 
to predict the probability distribution of the answer
as follows:
\begin{equation}
    P(A|v,q) = c(f(v,q)),
\end{equation}
where $f(\cdot,\cdot)$ denotes the cross-attention fusion function
and $c(\cdot)$ denotes the classifier formed by two layers of MLP.
%

For the Egocentric VQA task, each question corresponds to a specific answer, 
and all answers in the dataset form a fixed set. 
Therefore, the model needs to predict one label from this set and can be optimized using the following cross-entropy loss function~\cite{mao2023cross}:
\begin{equation}
    L_{qa} = - \frac{1}{M} \sum_{(v,q)\in D} a_{gt}(v,q) \cdot \mathrm{log}(P(A|v,q)),
\end{equation}
where 
M denotes the number of samples in the dataset $D$ and $a_{gt}(v,q)$ represents the corresponding one-hot ground-truth.

\subsection{Counterfactual Sample Construction}


To enhance the model's ability to understand multiple events 
in egocentric VideoQA, we propose 
Event Description Paraphrasing (EDP)
strategy to construct positive and negative question samples. 
Specifically, for each question, 
we first employ SpaCy Event Detection, which utilizes SpyCy~\cite{jugran2021extractive} to extract event-related descriptions, such as "open the microwave" and "the first action".
%
To construct
positive questions, 
EDP replaces the verb or noun descriptions of the events in the questions with synonyms,
preserving the original semantic intent while introducing linguistic diversity. 
%
As shown in Figure~\ref{fig:3},
we replace the verb in the description "open the microwave" with its synonym (i.e., "turn on") by using WordNet in NTLK~\cite{schmitt2019replicable}.
%
Similarly, for some questions containing references like "the first action", the term "action" is replaced with "operation".
This strategy can help 
encourage the model to generalize across different event descriptions
with consistent semantics. 
For negative samples, 
%
EDP removes critical information from questions to enhance the model’s ability to focus on events and ignore irrelevant details under the constraint of contrastive loss. 
%
Specifically, we mask event-related descriptions
from the original questions 
with a special token "[MASK]" 
and alter temporal contexts by swapping terms like 
"before" with "after", as shown in figure~\ref{fig:3}. 
%
%
These perturbations enhance the model’s ability to focus on essential event cues. 
The generation of positive and negative samples is expressed as:
\begin{equation}
    Q^+ = SS(Q, SpaCy(Q), WordNet),
\end{equation}
\begin{equation}
    Q^- = TDC(EM(Q, SpaCy(Q), [Mask])),
\end{equation}
where $Q$ is the original question, $SS$ represents synonym substitution, and $EM$, $TDC$ represent event masking and time description conversion. $SpaCy(\cdot)$ denotes SpaCy event detection.

%
 
%
To 
enhance the focus on hand-object interactions in egocentric videos, 
we propose Core Interaction Mining (CIM)
strategy 
to construct positive and negative video samples to sharpen the model’s attention on critical action events. 
In egocentric videos, 
hand-object interactions dominate most frames, with key actions 
concentrated in the central view. 
%
This inspires CIM to first select the interactive part of each frame by Interaction Region Selection.
The positive sample is constructed by retaining the selected region, where essential interactions are most likely to occur, and masking the other regions.
%
Conversely, negative samples are generated by 
masking the selected region and retaining the surrounding areas, which often contain less relevant or distracting information. 
CIM strategy trains the model to distinguish significant visual cues from irrelevant ones.
%
We explore various extraction methods to obtain the central area containing the hand-object interaction. 
The specific methods will be detailed in the experimental section.
%
%
These operations of constructing positive and negative videos are expressed as:
\begin{equation}
    V^+ = Rt(V,IRS(V),region),\quad  V^- = Ms(V,IRS(V),region),
\end{equation}
where $V$ is the original video, 
$Rt$ denotes the retaining
operation, and $Ms$ denotes the masking operation.
$IRS(\cdot)$ denotes interaction region selection, and $region$ denotes the area selected by $IRS(\cdot)$.
%

{
As shown in Figure~\ref{fig:3}, based on the above two strategies, we generate corresponding positive sample pairs $(V^+, Q^+)$ and negative sample pairs $(V^-, Q^-)$ for each input sample $(V, Q)$.
Finally, we leverage these positive factual and negative counterfactual samples
for Counterfactual Sample-involved Contrastive Optimization.

\begin{table*}[!ht]
    \centering
    \caption{Performances on EgoTaskQA \textit{normal} split. Model performance is evaluated based on scope, type, semantics, and overall category. The prediction accuracy (\%) of the baseline and two-stage trained DMC$^3$ are shown in the last two columns of the table.
    The best and second-best results are indicated in bold+underline and bold respectively.
    }
    \begin{tabular}{c c c c c c c c | c c}
    \hline
        ~ & Category  & HGA~\cite{jiang2020reasoning} & HCRN~\cite{le2021hierarchical} & 
        EgoVLP~\cite{lin2022egocentric} & EgoVLPv2~\cite{pramanick2023egovlpv2} & VideoDistill~\cite{zou2024language} & MFAS~\cite{zhangmulti} & Baseline & DMC$^3$  \\ \hline
        \multirow{3}*{\rotatebox{90}{Scope}} & world  & 38.82 & 44.27 & 
        45.35 & 50.25 & 47.32 & \textbf{52.96} & 50.17&  \underline{\textbf{53.17}}\\ 
        ~ & intent  & 42.12 & 49.77 & 
        50.41 & 53.69 & 52.53 & 56.00 & \textbf{57.48}&  \underline{\textbf{58.33}}\\ 
        ~ & multi-agent  & 23.43 & 31.36 & 
        31.90 & 40.64 & 36.95 & \underline{\textbf{43.45}} & 40.83&  \textbf{42.05}\\ \hline
        \multirow{4}*{\rotatebox{90}{Type}} & descriptive  & 38.04 & 43.48 & 
        46.12 & 52.19 & 47.20 & \underline{\textbf{54.85}} & 51.00&  \textbf{52.58}\\ 
        ~ & predictive  & 25.57 & 36.56 & 
        38.91 & 41.41 & 40.43 & 44.71 & \textbf{51.26}&  \underline{\textbf{53.97}}\\ 
        ~ & counterfactual  & 41.94 & 48.00 & 
        44.47 & 48.16 & 49.64 & \textbf{51.45} &  50.54& \underline{\textbf{52.05}}\\ 
        ~ & explanatory  & 35.97 & 40.60 & 
        40.22 & 42.36 & 42.53 & \underline{\textbf{44.35}} &  42.50& \textbf{43.73}\\ \hline
        \multirow{4}*{\rotatebox{90}{Semantic}} & action  & 15.08 & 14.92 & 
        15.96 & 16.80 & \textbf{16.35} & \underline{\textbf{17.34}} & 15.73&  15.86\\ 
        ~ & object  & 19.09 & 45.31 & 
        51.47 & 63.87 & 54.64 & \underline{\textbf{70.25}} &  66.21& \textbf{66.70}\\ 
        ~ & state  & 55.65 & 68.28 & 
        64.02 & 70.90 & 72.37 & \textbf{76.37} &  76.36& \underline{\textbf{79.75}}\\ 
        ~ & change  & 68.38 & 67.38 & 
        69.14 & 72.87 & 71.47 & 73.88 &  \textbf{75.49}& \underline{\textbf{76.26}}\\ \hline
        \multirow{3}*{\rotatebox{90}{Overall}} & open  & 22.75 & 30.23 & 
        31.69 & 35.56 & -- & \underline{\textbf{38.95}} &  35.22& \textbf{36.75}\\ 
        ~ & binary  & 68.53 & 69.42 & 
        71.26 & 75.60 & -- & 75.86 & \textbf{76.69}&  \underline{\textbf{78.56}}\\ 
        ~ & all & 36.77 & 42.20 & 
        42.51 & 46.26 & 45.02 & 48.69 & \textbf{48.86} & \underline{\textbf{52.51}}\\ \hline
    \end{tabular}
    
    \label{tab:1}
\end{table*}


\subsection{Counterfactual Sample-involved Contrastive Optimization}

It is worth noting that our model is trained in two stages.
%
In the first stage, 
we only optimize the egocentric videoqa baseline with the cross-entropy loss $L_{qa}$ introduced in formula 2.
%
In the second stage, we 
perform counterfactual sample-involved contrastive optimization
to 
improve the answer prediction procedure as follows:
\begin{equation}
    Loss = L_{qa}+\alpha L_{pos}+\beta L_{neg}+\lambda L_{con},
\end{equation}
where $\alpha$, $\beta$ and $\lambda$ are hyper-parameters.

\begin{figure}
    \centering
    \includegraphics[width=1.0\linewidth]{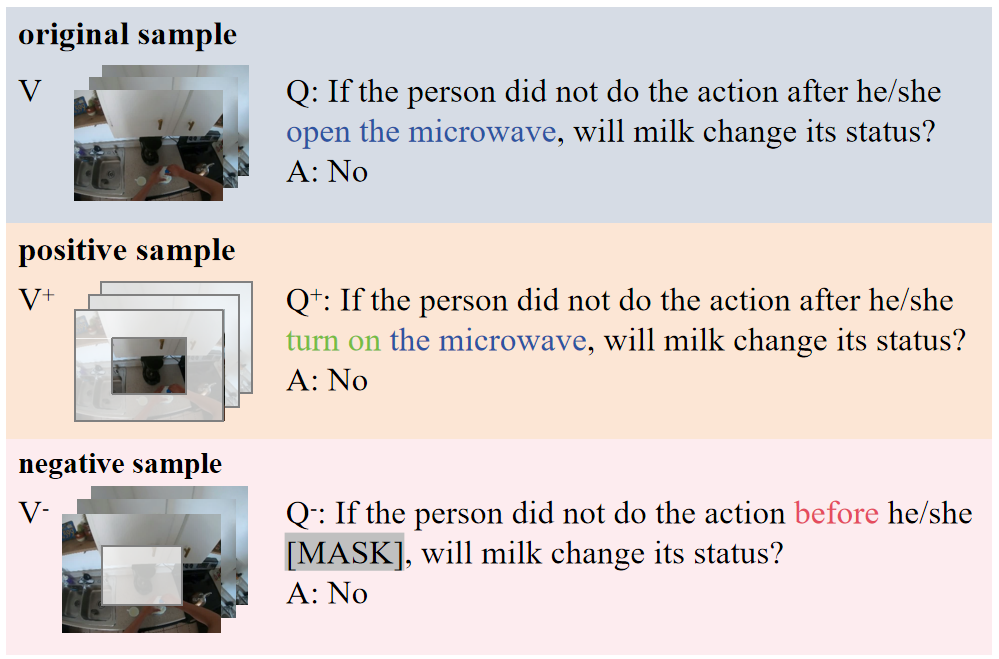}
    \caption{Examples of positive factual and negative counterfactual samples derived from an original sample. 
    %
    The white part of the video frame is the mask area. The event in the text is replaced with a special token "[MASK]".
}
    \label{fig:3}
\end{figure}

%
The cross-entropy loss functions $L_{pos}$ and $L_{neg}$ are 
designed to optimize the prediction results for positive and negative samples, respectively.
Specifically, 
we first feed the constructed
positive samples $(V^+,Q^+)$ and negative samples $(V^-,Q^-)$ 
into the baseline to predict the answers as follows:
\begin{equation}
    P(A^+|v^+,q^+) = c(f(v^+,q^+)),\quad P(A^-|v^-,q^-) = c(f(v^-,q^-)),
\end{equation}
where $(v^+,q^+)$ and $(v^-,q^-)$ are the encoded features of positive and negative samples respectively.
%
%
Then, we leverage 
the cross entropy loss function similar to Formula 2 to obtain $L_{pos}$ and $L_{neg}$
, helping reinforcing the fusion of the two modal features based on factual and counterfactual samples. 

Finally, for the answer distribution $P(A|v,q)$ of the 
original sample, abbreviated as $P_{A}$,
we aim to align it with positive samples and keep it away from negative samples.
Therefore, we treat $P_{A}$ as an anchor and propose the following contrastive loss resembles the InfoNCE loss~\cite{oord2018representation}.
\begin{equation}
    L_{con} = - \mathrm{log}\frac{exp(s(P_{A},P_{A^+})/\tau)}{exp(s(P_{A},P_{A^+})/\tau)+exp(P_{A},P_{A^-})/\tau)},
\end{equation}
where $s(p,q)=\frac{p^\mathsf{T}q}{||p||\cdot||q||}$ denotes the dot product between $l_2$ normalized $p$ and $q$. $||\cdot||$ represents $l_2$ normalization, and $\tau$ is a scalar temperature parameter.

By integrating EDP and CIM with contrastive learning, 
we improve the model’s ability to understand multiple events and focus on hand-object interactions.
Under the constraint of the counterfactual contrastive loss,
we can align the original question embeddings with the positive samples that retain event semantics, while distancing them from the negative samples that obscure event details. 
This enables the model to recognize consistent event patterns despite linguistic variations, fostering robust multi-event comprehension.
%
Furthermore, it helps minimize the distance between 
the
original video embeddings and the positive samples that highlight interactive regions, 
while repelling negative samples with irrelevant visuals. 
This sharpens the model’s focus on critical hand-object interactions, enhancing action-centric reasoning.

\section{Experiments}
In this section, we first introduce the datasets and 
the experimental setup.
%
Then, we show the experimental results on the \textit{normal} and \textit{indirect} splits of EgoTaskQA~\cite{jia2022egotaskqa} and QAEGO4D~\cite{barmann2022did}, and compare them with other methods. 
%
Finally, 
we conduct comprehensive ablation studies to verify the 
effectiveness of the proposed strategies.

\subsection{Datasets}
To evaluate the effectiveness of our proposed DMC$^3$,
we conduct extensive experiments on EgoTaskQA~\cite{jia2022egotaskqa} and QAEGO4D~\cite{barmann2022did} datasets. 

EgoTaskQA~\cite{jia2022egotaskqa} 
dataset contains 40k balanced question-answer pairs corresponding to 2k egocentric videos. 
These videos are selected from the LEMMA dataset~\cite{jia2020lemma}, with an average length of 36.9 seconds. 
EgoTaskQA has a detailed categorization that encompasses four dimensions: scope, type, semantic, and overall. 
%
%
%
%
%
It is segmented into two subsets: \textit{normal} and \textit{indirect}. 
The \textit{normal} split involves randomly selecting questions according to the answer distribution. 
In contrast, the \textit{indirect} subset consists of questions strongly tied to actions and objects, demanding more in-depth reasoning.

%
QAEGO4D~\cite{barmann2022did} is built upon the Ego4D~\cite{grauman2022ego4d} dataset and features a substantial collection of egocentric videos, each up to 8 minutes long, accompanied by natural language questions and answers, as well as corresponding time window annotations.
It contains a total of 1,325 videos and 14,513 questions, and the size of the answer set is 4,837.
The data collection process relies on manual annotations, where annotators view video clips based on the provided time windows to ensure the accuracy and relevance of the answers. This dataset is specifically designed for Egocentric VideoQA, focusing on the efficient retrieval of information from long egocentric videos.

\subsection{Experimental Setup}
\textbf{Implementation Details.} In our experiments, 
we randomly sample 16 frames from the video and feed them into the video encoder, TimeSformer~\cite{bertasius2021space}, which has 12 layers and is configured with the patch size of 16 $\times$ 16.
The encoded visual features have 3 channels, each with a height and width of 224 pixels. The text encoder is a 12-layer Transformer~\cite{han2022survey}. It generates a 512-dimensional embedding for each word, with the sentence length set to a fixed value of 30.

The training process of our model is divided into two stages. 
In the first stage, we focus exclusively on training the baseline model
, isolating it from other components 
to establish a strong foundational representation.
we incorporate pre-trained parameters from EgoVLPv2~\cite{pramanick2023egovlpv2}, a model pre-trained on egocentric video-language pairs, which we fine-tune to adapt to our specific task. This fine-tuning runs for 35 epochs with a batch size of 32.
In the second stage, all modules are trained simultaneously for 5 epochs.
We utilize Adam~\cite{kingma2014adam} optimizer with a weight decay of 0.0001, and set the temperature $\tau$ in Equation 5 to 0.1. The trade-off parameters $\alpha$, $\beta$, and $\lambda$ in Equation 6 are set to 0.8, 1, and 0.01, respectively.

\textbf{Evaluation Protocol.}
In this work, we primarily focus on prediction accuracy as the key metric.
%
Additionally, we employ machine translation metrics, 
METEOR~\cite{banerjee2004meteor} and ROUGE~\cite{lin2004rouge}, to assess the quality of responses in QAEGO4D~\cite{barmann2022did}. 
%
%
METEOR considers both precision and recall, incorporating synonymy and stemming to evaluate semantic similarity. 
Finally, ROUGE focuses on recall by comparing the overlap of n-grams between the generated response and ground truth. 
Together, these metrics offer a comprehensive evaluation framework for assessing the performance of our model, with higher scores indicating better performance.

\begin{table*}[!ht]
    \centering
    \caption{Performances on EgoTaskQA \textit{indirect} split.
    Model performance is evaluated based on scope, type, semantics, and overall category. The prediction accuracy (\%) of the baseline and two-stage trained DMC$^3$ are shown in the last two columns of the table. The best and second-best results are indicated in bold+underline and bold respectively.}
    \begin{tabular}{c c  c c c c c c | c c}
    \hline
        ~ & Category  & HGA~\cite{jiang2020reasoning} & HCRN~\cite{le2021hierarchical} & 
        EgoVLP~\cite{lin2022egocentric} & EgoVLPv2~\cite{pramanick2023egovlpv2} & VideoDistill~\cite{zou2024language} & MFAS~\cite{zhangmulti} & Baseline & DMC$^3$  \\ \hline
        \multirow{3}*{\rotatebox{90}{Scope}} & world  & 31.29 & 44.04 & 
        41.45 & 44.90 & 47.82 & \textbf{48.62} & 47.49 & \underline{\textbf{49.08}}   \\
        ~ & intent  & 20.42 & 47.02 & 
        33.61 & 40.48 &\textbf{ 49.61} & 42.55  &  47.22 & \underline{\textbf{52.95}} \\ 
        ~ & multi-agent  & 17.74 & 30.11 & 
        29.06 & 32.24 & \textbf{35.04} & 31.21  &  34.48 & \underline{\textbf{36.32}} \\ \hline
        \multirow{4}*{\rotatebox{90}{Type}} & descriptive  & 29.01 & 42.02 & 
        40.30 & 45.84 & 45.13 & \underline{\textbf{48.79}}  &  46.03 & \textbf{48.29} \\
        ~ & predictive  & 15.16 & 46.32 & 
        22.61 & 43.69 & \textbf{52.83} & 46.74  &  51.57 & \underline{\textbf{56.93}} \\
        ~ & counterfactual  & 33.01 & 43.64 & 
        37.70 & 38.94 & \underline{\textbf{43.97}} & 41.43  &  41.08 & \textbf{43.72} \\ 
        ~ & explanatory  & 24.00 & 39.69 & 
        35.91 & 39.10 & \underline{\textbf{43.75}} & 39.06  &  41.98 & \textbf{43.72} \\ \hline
        \multirow{4}*{\rotatebox{90}{Semantic}} & action  & 26.15 & 29.61 & 
        29.71 & 29.09 & \underline{\textbf{30.34}} & \textbf{30.13}  &  29.52 & 29.99 \\ 
        ~ & object  & 7.02 & 32.20 & 
        32.94 & 40.19 & \textbf{45.97} & 44.11  &  44.48 & \underline{\textbf{47.10}} \\ 
        ~ & state  & 17.67 & 41.81 & 
        36.52 & 41.69 & \textbf{49.77} & 44.63  &  49.15 & \underline{\textbf{53.66}} \\ 
        ~ & change  & 47.22 & 56.27 & 
        51.84 & 56.38 & 53.98 & \underline{\textbf{60.78}}  &  54.08 & \textbf{56.15} \\ \hline
        \multirow{3}*{\rotatebox{90}{Overall}} & open  & 8.66 & 27.82 & 
        27.04 & 29.14 & -- & 32.44  &  \textbf{32.82} & \underline{\textbf{35.34}} \\ 
        ~ & binary  & 53.72 & 59.29 & 
        55.28 & 59.68 & -- & \underline{\textbf{63.02}}  &  59.01 & \textbf{60.34} \\ 
        ~ & all & 28.36 & 41.56 & 
        38.69 & 42.28 & 44.77 & \textbf{45.40} & 44.17 & \underline{\textbf{46.04}} \\ \hline
    \end{tabular}
    
    \label{tab:2}
\end{table*}

\begin{table}[!ht]
    \centering
    \caption{Performances on QAEGO4D. Among them, Acc(\%) represents the prediction accuracy, and 
    METEOR, ROUGE are indicators of machine translation. }
    \begin{tabular}{l|lccc}
    \hline
        Methods & Acc  & METEOR & ROUGE  \\ \hline
        HCRN~\cite{le2021hierarchical} & 10.3   & 17.2 & 25.7  \\  
        CMCIR~\cite{liu2023cross} & 9.7& 16.5& 24.7\\ 
        EgoVLP~\cite{lin2022egocentric} & 10.2& 17.0& 25.4\\ 
        EgoVLPv2~\cite{pramanick2023egovlpv2} & 10.3&  17.4& 25.8\\  
        MFAS~\cite{zhangmulti} &  \underline{12.7}& \underline{18.3}& \underline{27.0}\\ \hline
        baseline &11.4 &  17.6& 25.4\\ 
        DMC$^3$ &\textbf{13.2}  & \textbf{18.4}& \textbf{27.5}\\  \hline
    \end{tabular}
    
    \label{tab:3}
\end{table}

\subsection{Quantitative Results}
In our comparison on the EgoTaskQA dataset, we evaluate against various methods including HGA~\cite{jiang2020reasoning}, HCRN~\cite{le2021hierarchical}, EgoVLP~\cite{lin2022egocentric}, EgoVLPv2~\cite{pramanick2023egovlpv2}, VideoDistill~\cite{zou2024language}, and MFAS~\cite{zhangmulti}. 

%

%
Table~\ref{tab:1} presents the comparison of our results with the aforementioned methods on the EgoTaskQA 
\textit{normal} dataset.
The outcomes of baseline and the final results after two-stage training are displayed in the last two columns. 
In each category, our model surpasses the baseline, achieving the highest prediction accuracy of 52.51\% across the entire dataset.
In addition, We achieve the highest prediction accuracy in several categories, including word and intent within the scope dimension, as well as predictive and counterfactual types. 
We attained the second highest accuracy in various other categories, except for action. 
When comparing our second highest accuracy with the best results from other models, our performance remains competitive, with only a small gap between our results and the top predictions
Therefore, our method demonstrates strong overall performance.

In Table~\ref{tab:2}, the results of our approach on the EgoTaskQA \textit{indirect} dataset are showcased. 
Following the first stage of training, our baseline achieves an overall prediction accuracy of 44.17\% after pre-training with the EgoVLPv2~\cite{pramanick2023egovlpv2} visual encoder weights. 
For most categories, the prediction accuracy of our baseline exceeds that of EgoVLPv2. 
After the second stage of training, DMC$^3$ achieves an accuracy of 46.04\%, establishing a state-of-the-art performance. 
%
%
By comparing the details of the two tables, we observe that our method achieves the 
highest prediction accuracy 
with respective to the category of all in
both \textit{normal} and \textit{indirect} settings,
%
and ranks either first or second in most categories, showing only a small difference from the best results.
%
At the same time, we also observe that our improvement in the action category was not very pronounced. 
Upon examining the dataset, we find that this limitation is due to the insufficient event descriptions related to this type of question~\cite{jia2022egotaskqa}.
This indicates that our method is effective in addressing the specific challenges associated with Egocentric VideoQA.
%

%
To further validate the performance of our model, we conduct experiments on the QAEGO4D dataset~\cite{barmann2022did}, as summarized in Table~\ref{tab:3}.
%
%
After two stages of training, our DMC$^3$ model achieves an accuracy of 13.2\%, surpassing the previous best performing model, MFAS~\cite{zhangmulti}, and exceeding the baseline by 1.8\%.
In terms of machine translation metrics, DMC$^3$ 
%
ranks first with scores of 18.4 on Meteor and 27.5 on Rouge, respectively.

\subsection{Ablation Studies}
We conduct ablation studies on the counterfactual sample construction module and hyper-parameters using the \textit{indirect} split of EgoTaskQA~\cite{jia2022egotaskqa}.
In order to verify the effectiveness of counterfactual sample construction module, we list several construction methods of the two modalities in Table~\ref{tab:4} and present the experimental results in Table~\ref{tab:5}.
Among them, $f_q$ and $f_v$ mean that the original samples remain unchanged, and those with digital identifiers represent their corresponding construction operations.
%
%
%
To ensure fairness, when positive and negative samples are generated for only one modality, they are input into the baseline alongside the original samples of the other modality. 
At this point, 
the loss function is modified to $Loss = L_{qa} + \lambda L_{con}$, where the $\lambda$ is set to 0.01.

\textbf{Impact of Event Description Paraphrasing Methods. }
In the textual modality, we evaluate three sample generation methods for Event Description Paraphrasing.
All three methods generate positive samples through synonym substitution.
However, they differ in how they create negative samples: $f_{q1}$ masks the event, $f_{q2}$ alters the time description, and $f_{q3}$ applies both techniques simultaneously.
%
%
As shown in Table~\ref{tab:5}, the overall accuracy of $\{f_{q3},f_v\}$ reaches 44.18\%, %
outperforming both $\{f_{q},f_v\}$, $\{f_{q1},f_v\}$ and $\{f_{q2},f_v\}$, demonstrating its effectiveness in enhancing event-focused reasoning.

\begin{table*}[!t]
    \centering
    \caption{Methods of constructing positive and negative samples for visual and textual modalities are listed. 
    }
    \begin{tabular}{l|ll}
    \hline
        Variants & positive question $Q^+$ & negative question $Q^-$  \\ \hline
        $f_{q} $ & - & -  \\ 
        $f_{q1} $ & synonym substitution & mask event  \\ 
        $f_{q2} $ & synonym substitution & time description conversion  \\ 
        $f_{q3} $ & synonym substitution & mask event, time description conversion \\ \hline
        Variants & positive video $V^+$ & negative video $V^-$  \\ \hline
        $f_{v} $ & - & -  \\ 
        $f_{v1} $  & Retain center 1/4 of each frame& Mask center 1/4 of each frame  \\ 
        $f_{v2} $  & Retain lower middle 1/4 of each frame & Mask lower middle 1/4 of each frame  \\ 
        $f_{v3} $  & Retain lower middle 3/8 of each frame & Mask lower middle 3/8 of each frame  \\ 
        $f_{v4} $  & Retain hand-object regions & Mask hand-object regions  \\ \hline
    \end{tabular}
    \label{tab:4}
\end{table*}

\begin{table}[!t]
    \centering
    \caption{Ablation studies of different methods for counterfactual samples construction module on the \textit{indirect} split of EgoTashQA. The numbers correspond to the construction method in Table 4.}
    \begin{tabular}{ll|ccc}
    \hline
        \multicolumn{2}{l|}{Modify}  & Open & Binary & All \\ \hline
        $f_{q}$ & $f_v$& 32.70 & 59.02 & 44.11  \\  \hline
        $f_{q1}$ & $f_v$ & 33.26 & 57.54 & 44.09 \\
        $f_{q2}$ & $f_v$ & 33.42 & 57.68 & 44.12  \\ 
        $f_{q3}$ & $f_v$ & 33.50  & 58.15 & 44.18  \\  \hline
        $f_q$& $f_{v1}$ & 33.06 & 59.36 & 44.46  \\ 
        $f_q$ & $f_{v2}$ & 32.10 & 60.27 & 44.31  \\ 
        $f_q$ & $f_{v3}$ & 32.54 & 58.75 & 43.99  \\ 
        $f_q$ & $f_{v4}$ & 33.51 & 58.11 & 43.79  \\  \hline
        $f_{q3}$ & $f_{v1}$ & 33.69 & 60.29 & 45.22  \\ 
        $f_{q3}$ & $f_{v1}+f_{v4}$ & 33.94 & 59.97 & 45.23 \\  \hline
    \end{tabular}
    
    \label{tab:5}
\end{table}

\textbf{Impact of Core Interaction Mining Methods. }
In the visual modality, we investigate various strategies of Core Interaction Mining. 
Specifically, we compare three fixed region selection methods and a dynamic selection that incorporates hand object detection~\cite{Shan20}.
%
$f_{v1}$, $f_{v2}$, and $f_{v3}$ 
respectively represent the central 1/4 region, the lower middle 1/4 region, and the lower middle 3/8 region retained in the positive sample, and masked in the negative sample.
Meanwhile, $f_{v4}$ refers to the dynamic area detected by the pre-trained 
hand object detection~\cite{Shan20} in each frame as the counterfactual sample.
Our comparison reveals that the prediction accuracy of $\{f_{q},f_{v1}\}$ and $\{f_{q},f_{v2}\}$ across the entire set is higher than $\{f_{q},f_{v}\}$, while $\{f_{q},f_{v3}\}$ and $\{f_{q},f_{v4}\}$ fall short. 
Among them, $\{f_{q},f_{v1}\}$ achieves the highest accuracy of 44.46\% overall, surpassing other approaches. 
The superior performance of $\{f_{q},f_{v1}\}$ and $\{f_{q},f_{v2}\}$ suggests that focusing on a specific range of regions enhances predictive capability. 
In contrast, the reduced accuracy of $\{f_{q},f_{v3}\}$ stems from the larger retained region, which introduces noise. 
Similarly, the reduced accuracy of $\{f_{q},f_{v4}\}$ is attributed to dynamic detection yielding small areas, omitting critical background.

\textbf{Impact of combining EDP and CIM. }
%
%
We further explore sample construction methods that combine the best of both modalities and the role of hand object detection.
%
Initially, we combine the best-performing generation methods for each modality as $\{f_{q3},f_{v1}\}$, achieving an overall prediction accuracy of 45.22\%, which exceeds $\{f_{q},f_{v}\}$ across 
open, binary and all categories.
Next, $\{f_{q},f_{v1}+f_{v4}\}$ explores the impact of hand-object detection 
by adding $f_{v4}$, 
resulting in an accuracy of 45.23\%, which is comparable to that of the previous method.
%
This stability suggests that $f_{v1}$ captures most hand-object interactions and provides sufficient background context for the model. 
Additionally, our method is less computationally intensive than relying on the detector screening frame-by-frame.

\begin{figure}
    \centering
    \includegraphics[width=0.98\linewidth]{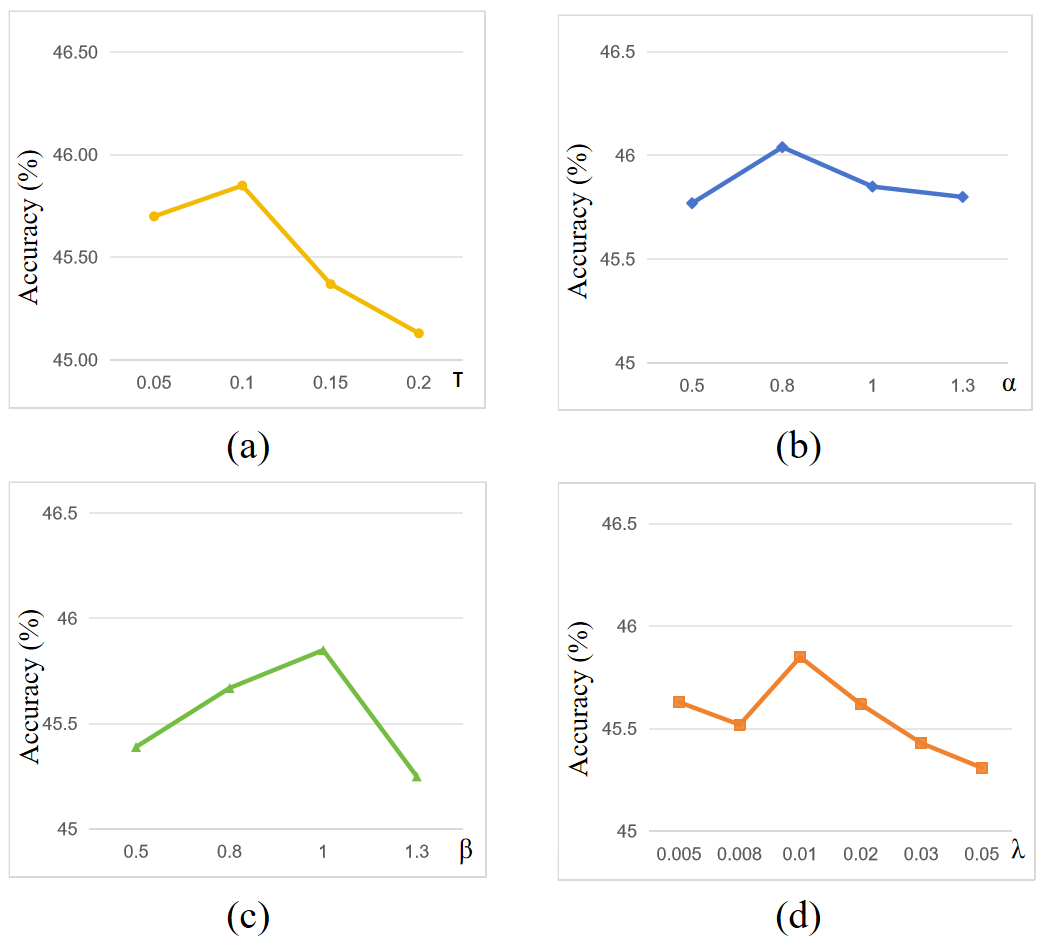}
    \caption{Ablation studies of hyper-parameters on the \textit{indirect} split of EgoTashQA. (a), (b), (c) and (d) are the performance changes of temperature parameter  $\tau$ and weight parameter $\alpha$, $\beta$, $\lambda$, respectively.
}
    \label{fig:4}
\end{figure}

\textbf{Impact of Hyper-parameters. }
We also conduct ablation experiments on the hyper-parameters, and the results are illustrated in Figure~\ref{fig:4}. 
For each parameter, we keep other parameters unchanged and set the initial \{$\tau$, $\alpha$, $\beta$, $\lambda$\} to \{0.1, 1, 1, 0.01\}.
The prediction accuracy is highest when the $\tau$ is 0.1.
The range of values for $\alpha$ is \{0.5, 0.8, 1\}. 
The highest prediction accuracy is attained when $\alpha$ is set to 0.8. 
The range of values for $\beta$  that of $\alpha$, and the most accurate prediction is achieved at $\beta=1$. 
For $\lambda$, the values ranges from \{0.005, 0.008, 0.01, 0.02, 0.03, 0.05\}. 
The optimal effect is observed with $\lambda$ set to 0.01, resulting in a prediction accuracy of 45.85\%. 
%
Finally, when $\alpha$, $\beta$, $\lambda$ are taken \{0.8, 1, 0.01\} separately, our model achieves the highest prediction accuracy.

\section{Conclusions}
In this paper, we propose the Dual-Modal Counterfactual Contrastive Construction model for Egocentric VideoQA, enhancing the understanding of hand-object interactions and multiple events. 
In the counterfactual sample construction module, we design Event Description Paraphrasing and Core Interaction Mining methods to generate positive and negative samples for both text and visual modalities. 
These samples are input into the baseline model alongside the original samples, with training optimized by contrastive loss in counterfactual sample-involved contrastive optimization. This ensures that the fused features corresponding to the original samples are close to the positive sample features and distant from the negative sample features.
Extensive experiments have validated the effectiveness of our proposed model, demonstrating its capability to learn high-quality representations for Egocentric VideoQA.


\begin{acks}
This work is supported by This work was supported by the National Natural Science Foundation of China under Grant 62325206, the Key Research and Development Program of Jiangsu Province under Grant BE2023016-4. 
This work was supported by National Natural Science
Foundation of China under Grants 62036012, U23A20387,
62322212, in part by Pengcheng Laboratory Research Project under Grant PCL2023A08, and also in part by the Postdoctoral Fellowship Program of CPSF under Grant Number GZC20251036.
\end{acks}

\bibliographystyle{ACM-Reference-Format}
\bibliography{sample-base}

@String{Computing = "Computing" }

@String{Computer = "{IEEE} Computer" }

@String{Springer = "Springer-Verlag" }

@INPROCEEDINGS{Shan20, 
    author = {Shan, Dandan and Geng, Jiaqi and Shu, Michelle  and Fouhey, David},
    title = {Understanding Human Hands in Contact at Internet Scale},
    booktitle = CVPR, 
    year = {2020} 
}

@article{majumdar2023we,
  title={Where are we in the search for an artificial visual cortex for embodied intelligence?},
  author={Majumdar, Arjun and Yadav, Karmesh and Arnaud, Sergio and Ma, Jason and Chen, Claire and Silwal, Sneha and Jain, Aryan and Berges, Vincent-Pierre and Wu, Tingfan and Vakil, Jay and others},
  journal={Advances in Neural Information Processing Systems},
  volume={36},
  pages={655--677},
  year={2023}
}

@article{phan2024henasy,
  title={HENASY: Learning to Assemble Scene-Entities for Interpretable Egocentric Video-Language Model},
  author={Phan, Khoa Vo Thinh and Tran, Kashu Yamazaki Minh and Le, Ngan},
  journal={Advances in Neural Information Processing Systems},
  year={2024}
}

@inproceedings{di2024grounded,
  title={Grounded Question-Answering in Long Egocentric Videos},
  author={Di, Shangzhe and Xie, Weidi},
  booktitle={Proceedings of the IEEE/CVF Conference on Computer Vision and Pattern Recognition},
  pages={12934--12943},
  year={2024}
}

@inproceedings{chen2024egocentric,
  title={Egocentric Vehicle Dense Video Captioning},
  author={Chen, Feiyu and Xu, Cong and Jia, Qi and Wang, Yihua and Liu, Yuhan and Zhang, Haotian and Wang, Endong},
  booktitle={Proceedings of the 32nd ACM International Conference on Multimedia},
  pages={137--146},
  year={2024}
}

@inproceedings{huang2024egoexolearn,
  title={EgoExoLearn: A Dataset for Bridging Asynchronous Ego-and Exo-centric View of Procedural Activities in Real World},
  author={Huang, Yifei and Chen, Guo and Xu, Jilan and Zhang, Mingfang and Yang, Lijin and Pei, Baoqi and Zhang, Hongjie and Dong, Lu and Wang, Yali and Wang, Limin and others},
  booktitle={Proceedings of the IEEE/CVF Conference on Computer Vision and Pattern Recognition},
  pages={22072--22086},
  year={2024}
}

@article{mu2024embodiedgpt,
  title={Embodiedgpt: Vision-language pre-training via embodied chain of thought},
  author={Mu, Yao and Zhang, Qinglong and Hu, Mengkang and Wang, Wenhai and Ding, Mingyu and Jin, Jun and Wang, Bin and Dai, Jifeng and Qiao, Yu and Luo, Ping},
  journal={Advances in Neural Information Processing Systems},
  volume={36},
  year={2024}
}

@article{plizzari2024outlook,
  title={An outlook into the future of egocentric vision},
  author={Plizzari, Chiara and Goletto, Gabriele and Furnari, Antonino and Bansal, Siddhant and Ragusa, Francesco and Farinella, Giovanni Maria and Damen, Dima and Tommasi, Tatiana},
  journal={International Journal of Computer Vision},
  pages={1--57},
  year={2024},
  publisher={Springer}
}

@inproceedings{fan2019egovqa,
  title={EgoVQA-an egocentric video question answering benchmark dataset},
  author={Fan, Chenyou},
  booktitle={Proceedings of the IEEE/CVF International Conference on Computer Vision Workshops},
  pages={0--0},
  year={2019}
}

@inproceedings{min2024morevqa,
  title={Morevqa: Exploring modular reasoning models for video question answering},
  author={Min, Juhong and Buch, Shyamal and Nagrani, Arsha and Cho, Minsu and Schmid, Cordelia},
  booktitle={Proceedings of the IEEE/CVF Conference on Computer Vision and Pattern Recognition},
  pages={13235--13245},
  year={2024}
}

@inproceedings{xiao2022video,
  title={Video as conditional graph hierarchy for multi-granular question answering},
  author={Xiao, Junbin and Yao, Angela and Liu, Zhiyuan and Li, Yicong and Ji, Wei and Chua, Tat-Seng},
  booktitle={Proceedings of the AAAI Conference on Artificial Intelligence},
  volume={36},
  number={3},
  pages={2804--2812},
  year={2022}
}

@inproceedings{jin2023knowledge,
  title={Knowledge-constrained answer generation for open-ended video question answering},
  author={Jin, Yao and Niu, Guocheng and Xiao, Xinyan and Zhang, Jian and Peng, Xi and Yu, Jun},
  booktitle={Proceedings of the AAAI Conference on Artificial Intelligence},
  volume={37},
  number={7},
  pages={8141--8149},
  year={2023}
}

@article{nagarajan2023egoenv,
  title={EgoEnv: Human-centric environment representations from egocentric video},
  author={Nagarajan, Tushar and Ramakrishnan, Santhosh Kumar and Desai, Ruta and Hillis, James and Grauman, Kristen},
  journal={Advances in Neural Information Processing Systems},
  volume={36},
  pages={60130--60143},
  year={2023}
}

@inproceedings{wang2023ego,
  title={Ego-only: Egocentric action detection without exocentric transferring},
  author={Wang, Huiyu and Singh, Mitesh Kumar and Torresani, Lorenzo},
  booktitle={Proceedings of the IEEE/CVF International Conference on Computer Vision},
  pages={5250--5261},
  year={2023}
}

@inproceedings{wang2023all,
  title={All in one: Exploring unified video-language pre-training},
  author={Wang, Jinpeng and Ge, Yixiao and Yan, Rui and Ge, Yuying and Lin, Kevin Qinghong and Tsutsui, Satoshi and Lin, Xudong and Cai, Guanyu and Wu, Jianping and Shan, Ying and others},
  booktitle={Proceedings of the IEEE/CVF Conference on Computer Vision and Pattern Recognition},
  pages={6598--6608},
  year={2023}
}

@article{banerjee2004meteor,
  title={Meteor: an automatic metric for mt evaluation with high levels of correlation with human judgments},
  author={Banerjee, Satanjeev and Lavie, Alon},
  journal={Proceedings of ACL-WMT},
  pages={65--72},
  year={2004}
}

@inproceedings{lin2004rouge,
  title={Rouge: A package for automatic evaluation of summaries},
  author={Lin, Chin-Yew},
  booktitle={Text summarization branches out},
  pages={74--81},
  year={2004}
}

@inproceedings{liu2023advancing,
  title={Advancing video question answering with a multi-modal and multi-layer question enhancement network},
  author={Liu, Meng and Zhang, Fenglei and Luo, Xin and Liu, Fan and Wei, Yinwei and Nie, Liqiang},
  booktitle={Proceedings of the 31st ACM International Conference on Multimedia},
  pages={3985--3993},
  year={2023}
}

@inproceedings{liu2023enhancing,
  title={Enhancing Vision-Language Pre-Training with Jointly Learned Questioner and Dense Captioner},
  author={Liu, Zikang and Chen, Sihan and Guo, Longteng and Li, Handong and He, Xingjian and Liu, Jing},
  booktitle={Proceedings of the 31st ACM International Conference on Multimedia},
  pages={5120--5131},
  year={2023}
}

@inproceedings{li2023vision,
  title={Do vision-language transformers exhibit visual commonsense? an empirical study of vcr},
  author={Li, Zhenyang and Guo, Yangyang and Wang, Kejie and Chen, Xiaolin and Nie, Liqiang and Kankanhalli, Mohan},
  booktitle={Proceedings of the 31st ACM International Conference on Multimedia},
  pages={5634--5644},
  year={2023}
}

@article{liu2023cross,
  title={Cross-modal causal relational reasoning for event-level visual question answering},
  author={Liu, Yang and Li, Guanbin and Lin, Liang},
  journal={IEEE Transactions on Pattern Analysis and Machine Intelligence},
  volume={45},
  number={10},
  pages={11624--11641},
  year={2023},
  publisher={IEEE}
}

@inproceedings{barmann2022did,
  title={Where did i leave my keys?-episodic-memory-based question answering on egocentric videos},
  author={B{\"a}rmann, Leonard and Waibel, Alex},
  booktitle={Proceedings of the IEEE/CVF Conference on Computer Vision and Pattern Recognition},
  pages={1560--1568},
  year={2022}
}

@article{jia2022egotaskqa,
  title={Egotaskqa: Understanding human tasks in egocentric videos},
  author={Jia, Baoxiong and Lei, Ting and Zhu, Song-Chun and Huang, Siyuan},
  journal={Advances in Neural Information Processing Systems},
  volume={35},
  pages={3343--3360},
  year={2022}
}

@article{damen2022epic,
  title={Epic-kitchens-100},
  author={Damen, Dima and Doughty, Hazel and Farinella, Giovanni Maria and Furnari, Antonino and Kazakos, Evangelos and Ma, Jian and Moltisanti, Davide and Munro, Jonathan and Perrett, Toby and Price, Will and others},
  journal={International Journal of Computer Vision},
  volume={130},
  pages={33--55},
  year={2022}
}

@inproceedings{grauman2022ego4d,
  title={Ego4d: Around the world in 3,000 hours of egocentric video},
  author={Grauman, Kristen and Westbury, Andrew and Byrne, Eugene and Chavis, Zachary and Furnari, Antonino and Girdhar, Rohit and Hamburger, Jackson and Jiang, Hao and Liu, Miao and Liu, Xingyu and others},
  booktitle={Proceedings of the IEEE/CVF Conference on Computer Vision and Pattern Recognition},
  pages={18995--19012},
  year={2022}
}

@article{pei2024egovideo,
  title={Egovideo: Exploring egocentric foundation model and downstream adaptation},
  author={Pei, Baoqi and Chen, Guo and Xu, Jilan and He, Yuping and Liu, Yicheng and Pan, Kanghua and Huang, Yifei and Wang, Yali and Lu, Tong and Wang, Limin and others},
  journal={arXiv preprint arXiv:2406.18070},
  year={2024}
}

@inproceedings{zhang2023helping,
  title={Helping hands: An object-aware ego-centric video recognition model},
  author={Zhang, Chuhan and Gupta, Ankush and Zisserman, Andrew},
  booktitle={Proceedings of the IEEE/CVF International Conference on Computer Vision},
  pages={13901--13912},
  year={2023}
}

@article{lin2022egocentric,
  title={Egocentric video-language pretraining},
  author={Lin, Kevin Qinghong and Wang, Jinpeng and Soldan, Mattia and Wray, Michael and Yan, Rui and Xu, Eric Z and Gao, Difei and Tu, Rong-Cheng and Zhao, Wenzhe and Kong, Weijie and others},
  journal={Advances in Neural Information Processing Systems},
  volume={35},
  pages={7575--7586},
  year={2022}
}

@inproceedings{zhao2023learning,
  title={Learning video representations from large language models},
  author={Zhao, Yue and Misra, Ishan and Kr{\"a}henb{\"u}hl, Philipp and Girdhar, Rohit},
  booktitle={Proceedings of the IEEE/CVF Conference on Computer Vision and Pattern Recognition},
  pages={6586--6597},
  year={2023}
}

@inproceedings{pramanick2023egovlpv2,
  title={Egovlpv2: Egocentric video-language pre-training with fusion in the backbone},
  author={Pramanick, Shraman and Song, Yale and Nag, Sayan and Lin, Kevin Qinghong and Shah, Hardik and Shou, Mike Zheng and Chellappa, Rama and Zhang, Pengchuan},
  booktitle={Proceedings of the IEEE/CVF International Conference on Computer Vision},
  pages={5285--5297},
  year={2023}
}

@inproceedings{zhangmulti,
  title={Multi-Factor Adaptive Vision Selection for Egocentric Video Question Answering},
  author={Zhang, Haoyu and Liu, Meng and Liu, Zixin and Song, Xuemeng and Wang, Yaowei and Nie, Liqiang},
  booktitle={Forty-first International Conference on Machine Learning},
  year={2024}
}

@inproceedings{zou2024language,
  title={Language-aware Visual Semantic Distillation for Video Question Answering},
  author={Zou, Bo and Yang, Chao and Qiao, Yu and Quan, Chengbin and Zhao, Youjian},
  booktitle={Proceedings of the IEEE/CVF Conference on Computer Vision and Pattern Recognition},
  pages={27113--27123},
  year={2024}
}

@inproceedings{jiang2020reasoning,
  title={Reasoning with heterogeneous graph alignment for video question answering},
  author={Jiang, Pin and Han, Yahong},
  booktitle={Proceedings of the AAAI Conference on Artificial Intelligence},
  volume={34},
  number={07},
  pages={11109--11116},
  year={2020}
}

@article{le2021hierarchical,
  title={Hierarchical conditional relation networks for multimodal video question answering},
  author={Le, Thao Minh and Le, Vuong and Venkatesh, Svetha and Tran, Truyen},
  journal={International Journal of Computer Vision},
  volume={129},
  number={11},
  pages={3027--3050},
  year={2021},
  publisher={Springer}
}

@inproceedings{lei2021less,
  title={Less is more: Clipbert for video-and-language learning via sparse sampling},
  author={Lei, Jie and Li, Linjie and Zhou, Luowei and Gan, Zhe and Berg, Tamara L and Bansal, Mohit and Liu, Jingjing},
  booktitle={Proceedings of the IEEE/CVF conference on computer vision and pattern recognition},
  pages={7331--7341},
  year={2021}
}

@inproceedings{chen2020counterfactual,
  title={Counterfactual samples synthesizing for robust visual question answering},
  author={Chen, Long and Yan, Xin and Xiao, Jun and Zhang, Hanwang and Pu, Shiliang and Zhuang, Yueting},
  booktitle={Proceedings of the IEEE/CVF conference on computer vision and pattern recognition},
  pages={10800--10809},
  year={2020}
}

@article{chen2023counterfactual,
  title={Counterfactual samples synthesizing and training for robust visual question answering},
  author={Chen, Long and Zheng, Yuhang and Niu, Yulei and Zhang, Hanwang and Xiao, Jun},
  journal={IEEE Transactions on Pattern Analysis and Machine Intelligence},
  volume={45},
  number={11},
  pages={13218--13234},
  year={2023},
  publisher={IEEE}
}

@inproceedings{zhang2021multi,
  title={Multi-level counterfactual contrast for visual commonsense reasoning},
  author={Zhang, Xi and Zhang, Feifei and Xu, Changsheng},
  booktitle={Proceedings of the 29th ACM International Conference on Multimedia},
  pages={1793--1802},
  year={2021}
}

@article{tian2020makes,
  title={What makes for good views for contrastive learning?},
  author={Tian, Yonglong and Sun, Chen and Poole, Ben and Krishnan, Dilip and Schmid, Cordelia and Isola, Phillip},
  journal={Advances in neural information processing systems},
  volume={33},
  pages={6827--6839},
  year={2020}
}

@inproceedings{bertasius2021space,
  title={Is space-time attention all you need for video understanding?},
  author={Bertasius, Gedas and Wang, Heng and Torresani, Lorenzo},
  booktitle={ICML},
  volume={2},
  number={3},
  pages={4},
  year={2021}
}

@article{oord2018representation,
  title={Representation learning with contrastive predictive coding},
  author={Oord, Aaron van den and Li, Yazhe and Vinyals, Oriol},
  journal={arXiv preprint arXiv:1807.03748},
  year={2018}
}

@article{han2022survey,
  title={A survey on vision transformer},
  author={Han, Kai and Wang, Yunhe and Chen, Hanting and Chen, Xinghao and Guo, Jianyuan and Liu, Zhenhua and Tang, Yehui and Xiao, An and Xu, Chunjing and Xu, Yixing and others},
  journal={IEEE transactions on pattern analysis and machine intelligence},
  volume={45},
  number={1},
  pages={87--110},
  year={2022},
  publisher={IEEE}
}

@inproceedings{mao2023cross,
  title={Cross-entropy loss functions: Theoretical analysis and applications},
  author={Mao, Anqi and Mohri, Mehryar and Zhong, Yutao},
  booktitle={International conference on Machine learning},
  pages={23803--23828},
  year={2023},
  organization={PMLR}
}

@inproceedings{jugran2021extractive,
  title={Extractive automatic text summarization using SpaCy in Python \& NLP},
  author={Jugran, Swaranjali and Kumar, Ashish and Tyagi, Bhupendra Singh and Anand, Vivek},
  booktitle={2021 International conference on advance computing and innovative technologies in engineering (ICACITE)},
  pages={582--585},
  year={2021},
  organization={IEEE}
}

@inproceedings{schmitt2019replicable,
  title={A replicable comparison study of NER software: StanfordNLP, NLTK, OpenNLP, SpaCy, Gate},
  author={Schmitt, Xavier and Kubler, Sylvain and Robert, J{\'e}r{\'e}my and Papadakis, Mike and LeTraon, Yves},
  booktitle={2019 sixth international conference on social networks analysis, management and security (SNAMS)},
  pages={338--343},
  year={2019},
  organization={IEEE}
}

@inproceedings{jia2020lemma,
  title={LEMMA: A Multi-view Dataset for LE arning M ulti-agent M ulti-task A ctivities},
  author={Jia, Baoxiong and Chen, Yixin and Huang, Siyuan and Zhu, Yixin and Zhu, Song-chun},
  booktitle={European Conference on Computer Vision},
  pages={767--786},
  year={2020},
  organization={Springer}
}

@article{kingma2014adam,
  title={Adam: A method for stochastic optimization},
  author={Kingma, Diederik P},
  journal={arXiv preprint arXiv:1412.6980},
  year={2014}
}

@ArtifactSoftware{R,
    title = {R: A Language and Environment for Statistical Computing},
    author = {{R Core Team}},
    organization = {R Foundation for Statistical Computing},
    address = {Vienna, Austria},
    year = {2019},
    url = {https://www.R-project.org/},
}

@misc{guo2020nonautoregressiveimagecaptioningcounterfactualscritical,
      title={Non-Autoregressive Image Captioning with Counterfactuals-Critical Multi-Agent Learning}, 
      author={Longteng Guo and Jing Liu and Xinxin Zhu and Xingjian He and Jie Jiang and Hanqing Lu},
      year={2020},
      eprint={2005.04690},
      archivePrefix={arXiv},
      primaryClass={cs.CL},
      url={https://arxiv.org/abs/2005.04690}, 
}

@inproceedings{wen-etal-2023-digging,
    title = "Digging out Discrimination Information from Generated Samples for Robust Visual Question Answering",
    author = "Wen, Zhiquan  and
      Wang, Yaowei  and
      Tan, Mingkui  and
      Wu, Qingyao  and
      Wu, Qi",
    editor = "Rogers, Anna  and
      Boyd-Graber, Jordan  and
      Okazaki, Naoaki",
    booktitle = "Findings of the Association for Computational Linguistics: ACL 2023",
    month = jul,
    year = "2023",
    address = "Toronto, Canada",
    publisher = "Association for Computational Linguistics",
    url = "https://aclanthology.org/2023.findings-acl.432/",
    doi = "10.18653/v1/2023.findings-acl.432",
    pages = "6910--6928"
}

@article{mashrur2024robust,
  title={Robust visual question answering via semantic cross modal augmentation},
  author={Mashrur, Akib and Luo, Wei and Zaidi, Nayyar A and Robles-Kelly, Antonio},
  journal={Computer Vision and Image Understanding},
  volume={238},
  pages={103862},
  year={2024},
  publisher={Elsevier}
}

@inproceedings{zhang2024if,
  title={What if the tv was off? examining counterfactual reasoning abilities of multi-modal language models},
  author={Zhang, Letian and Zhai, Xiaotong and Zhao, Zhongkai and Zong, Yongshuo and Wen, Xin and Zhao, Bingchen},
  booktitle={Proceedings of the IEEE/CVF Conference on Computer Vision and Pattern Recognition},
  pages={21853--21862},
  year={2024}
}

@article{ren2024embodied,
  title={Embodied intelligence toward future smart manufacturing in the era of AI foundation model},
  author={Ren, Lei and Dong, Jiabao and Liu, Shuai and Zhang, Lin and Wang, Lihui},
  journal={IEEE/ASME Transactions on Mechatronics},
  year={2024},
  publisher={IEEE}
}

@article{zhou2025egotextvqa,
  title={EgoTextVQA: Towards Egocentric Scene-Text Aware Video Question Answering},
  author={Zhou, Sheng and Xiao, Junbin and Li, Qingyun and Li, Yicong and Yang, Xun and Guo, Dan and Wang, Meng and Chua, Tat-Seng and Yao, Angela},
  journal={arXiv preprint arXiv:2502.07411},
  year={2025}
}

@article{dai2024gpt4ego,
  title={GPT4Ego: unleashing the potential of pre-trained models for zero-shot egocentric action recognition},
  author={Dai, Guangzhao and Shu, Xiangbo and Wu, Wenhao and Yan, Rui and Zhang, Jiachao},
  journal={IEEE Transactions on Multimedia},
  year={2024},
  publisher={IEEE}
}

@article{salehi2024actionatlas,
  title={ActionAtlas: A VideoQA Benchmark for Domain-specialized Action Recognition},
  author={Salehi, Mohammadreza Reza and Park, Jae Sung and Kusupati, Aditya and Krishna, Ranjay and Choi, Yejin and Hajishirzi, Hanna and Farhadi, Ali},
  journal={Advances in Neural Information Processing Systems},
  volume={37},
  pages={137372--137402},
  year={2024}
}

@inproceedings{wang2024videoagent,
  title={Videoagent: Long-form video understanding with large language model as agent},
  author={Wang, Xiaohan and Zhang, Yuhui and Zohar, Orr and Yeung-Levy, Serena},
  booktitle={European Conference on Computer Vision},
  pages={58--76},
  year={2024},
  organization={Springer}
}

@article{liu2023multimodal,
  title={Multimodal graph contrastive learning for multimedia-based recommendation},
  author={Liu, Kang and Xue, Feng and Guo, Dan and Sun, Peijie and Qian, Shengsheng and Hong, Richang},
  journal={IEEE Transactions on Multimedia},
  volume={25},
  pages={9343--9355},
  year={2023},
  publisher={IEEE}
}

@misc{chen2024groundedmultihopvideoqalongform,
      title={Grounded Multi-Hop VideoQA in Long-Form Egocentric Videos}, 
      author={Qirui Chen and Shangzhe Di and Weidi Xie},
      year={2024},
      eprint={2408.14469},
      archivePrefix={arXiv},
      primaryClass={cs.CV},
      url={https://arxiv.org/abs/2408.14469}, 
}

@article{wei2019eda,
  title={EDA: Easy data augmentation techniques for boosting performance on text classification tasks},
  author={Wei, Jason and Zou, Kai},
  journal={arXiv preprint arXiv:1901.11196},
  year={2019}
}

@inproceedings{falcon2022feature,
  title={A feature-space multimodal data augmentation technique for text-video retrieval},
  author={Falcon, Alex and Serra, Giuseppe and Lanz, Oswald},
  booktitle={Proceedings of the 30th ACM international conference on multimedia},
  pages={4385--4394},
  year={2022}
}

@INPROCEEDINGS{10888014,
  author={Zou, Jiayi and Jia, Gengyun and Bao, Bing-Kun},
  booktitle={ICASSP 2025 - 2025 IEEE International Conference on Acoustics, Speech and Signal Processing (ICASSP)}, 
  title={Causal Debiasing for Visual Commonsense Reasoning}, 
  year={2025},
  volume={},
  number={},
  pages={1-5},
  doi={10.1109/ICASSP49660.2025.10888014}}

@ARTICLE{10814063,
  author={Yuan, Mengqi and Jia, Gengyun and Bao, Bing-Kun},
  journal={IEEE Transactions on Multimedia}, 
  title={Relation Inference Enhancement Network for Visual Commonsense Reasoning}, 
  year={2025},
  volume={27},
  number={},
  pages={2221-2231},
  doi={10.1109/TMM.2024.3521725}}

@ARTICLE{10310122,
  author={Yuan, Mengqi and Jia, Gengyun and Bao, Bing-Kun},
  journal={IEEE Transactions on Multimedia}, 
  title={GPT-Based Knowledge Guiding Network for Commonsense Video Captioning}, 
  year={2024},
  volume={26},
  number={},
  pages={5147-5158},
  doi={10.1109/TMM.2023.3330070}}




\end{document}